\RequirePackage[svgnames]{xcolor}

\documentclass[11pt,letterpaper]{mystyle}

\usepackage[all]{hypcap}
\usepackage[svgnames]{xcolor}
\usepackage[comma,authoryear,compress]{natbib}
\usepackage{hyperref}[citecolor=lightblue]

\newcommand{\schoollogos}{%
  \noindent
  \makebox[\textwidth][l]{%
    \includegraphics[
      height=0.85cm,
      keepaspectratio
    ]{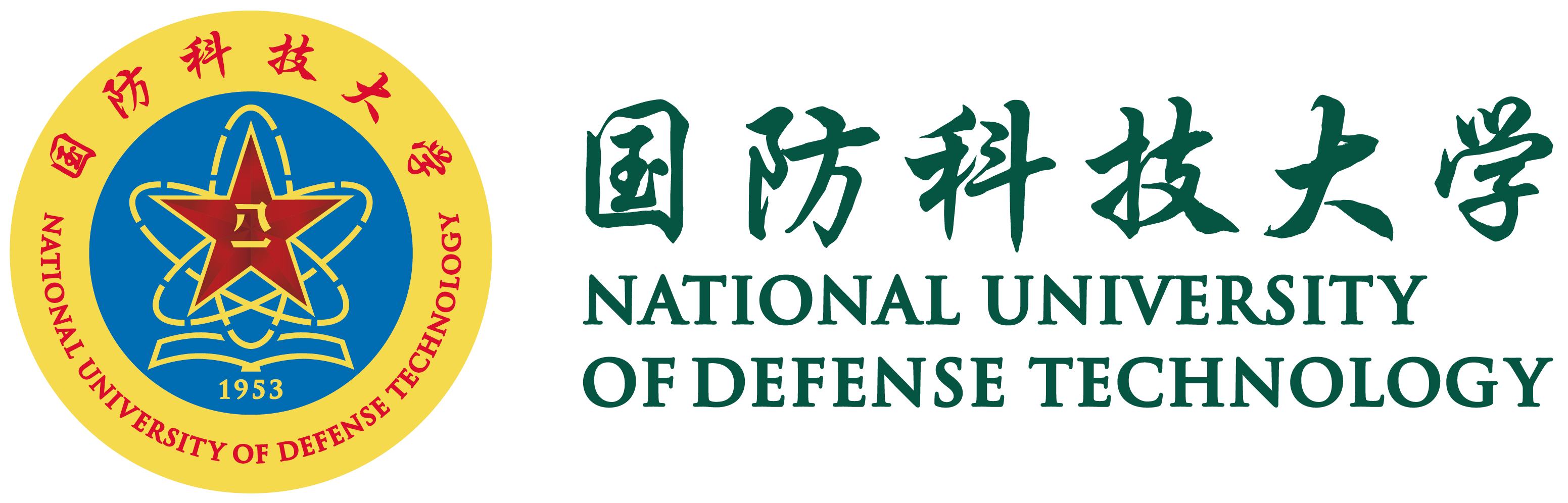}%
    \hspace{0.32cm}%
    \includegraphics[
      height=0.85cm,
      keepaspectratio
    ]{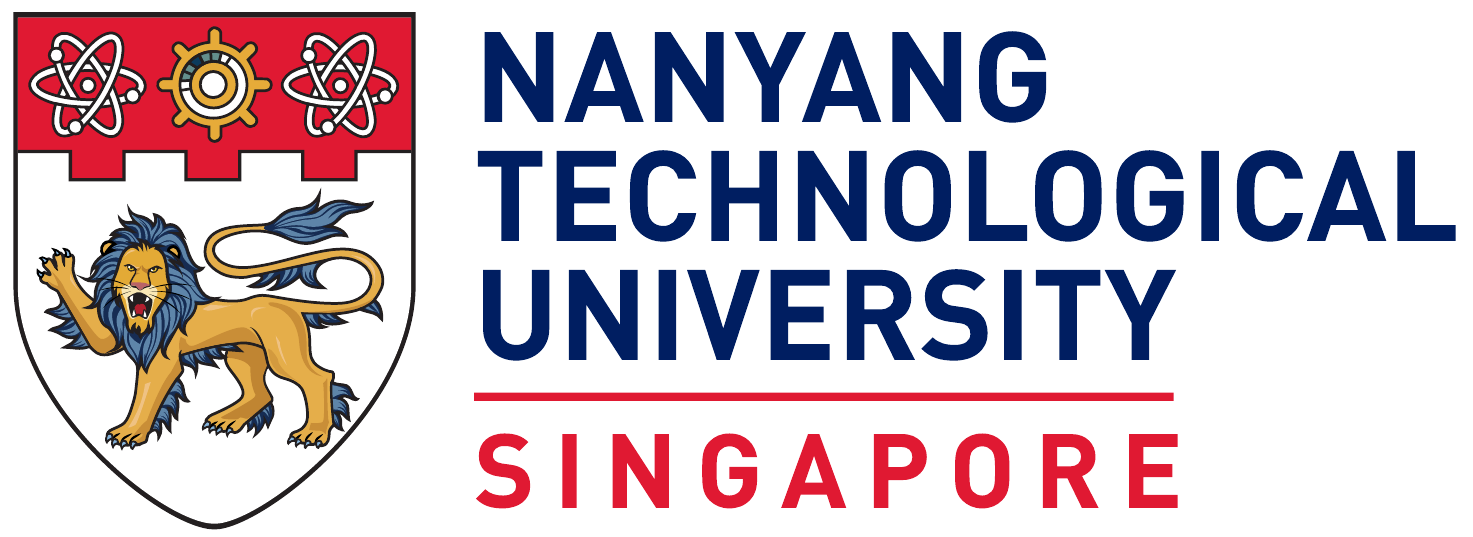}%
    \hspace{0.32cm}%
    \includegraphics[
      height=0.85cm,
      trim=220 700 220 750,
      clip
    ]{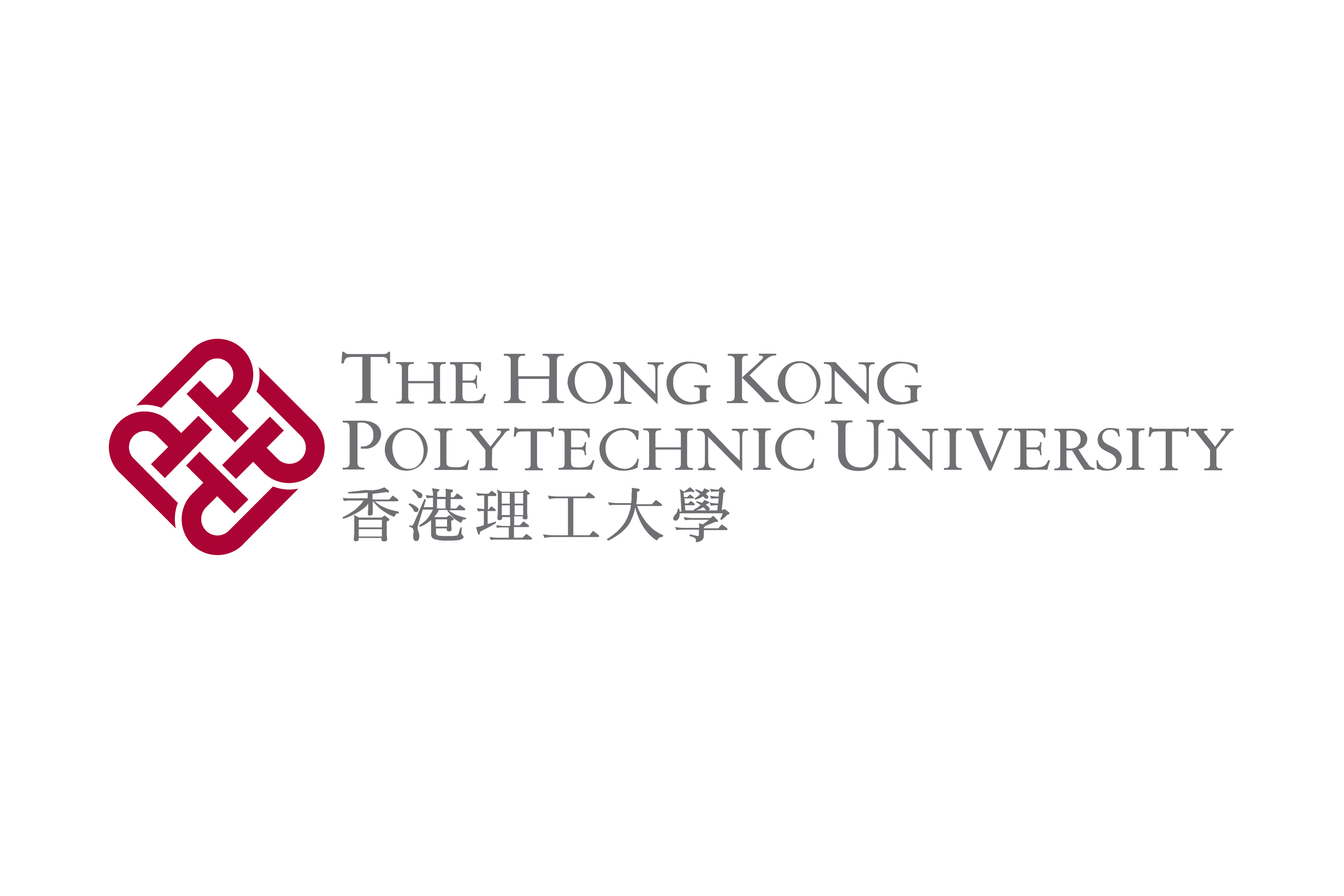}%
    \hspace{0.32cm}%
    \includegraphics[
      height=0.8cm,
      keepaspectratio
    ]{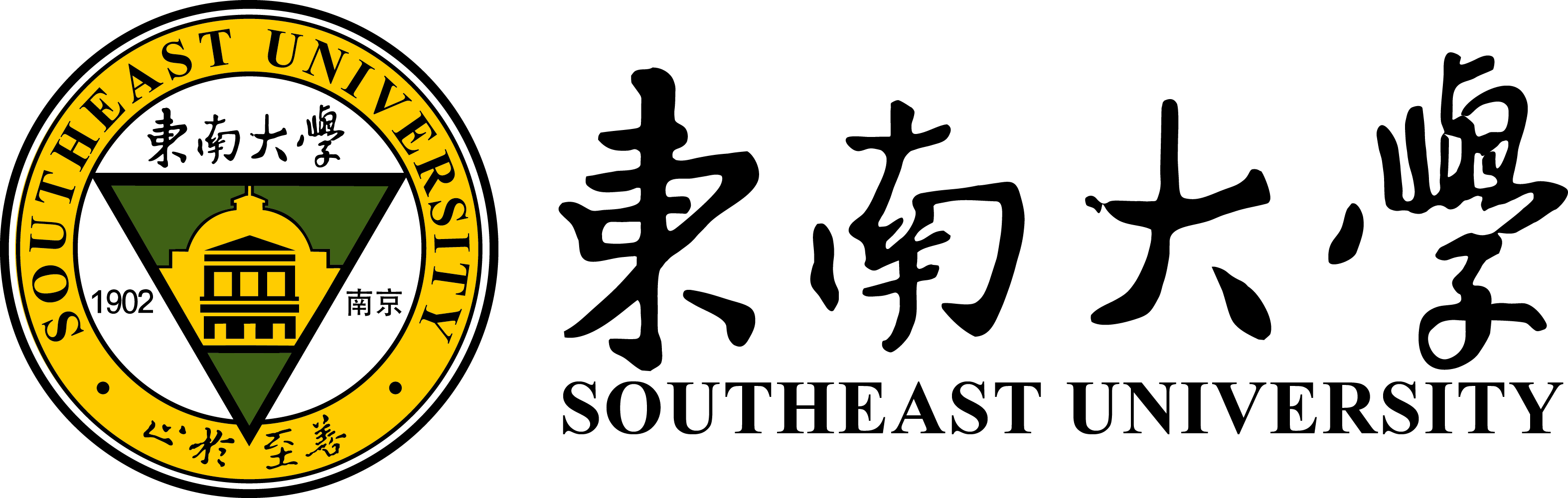}%
  }%
  \par\vspace{1.5mm}
  {\color{gray!30}\rule{\textwidth}{0.5pt}}
  \par\vspace{3mm}
}
\hypersetup{
    colorlinks = true,
    citecolor = {YaleBlue},
}

\usepackage{algorithm}
\usepackage{algorithmicx}
\usepackage{algpseudocode}
\usepackage{microtype}
\usepackage{graphicx}
\expandafter\def\csname ver@subfig.sty\endcsname{}
\usepackage{booktabs} %
\usepackage{float}
\usepackage{bigstrut}

\usepackage{amsmath}
\usepackage{amssymb}
\usepackage{mathtools}
\usepackage{amsthm}
\usepackage{mathrsfs}
\usepackage{nicefrac}
\usepackage{dsfont}
\usepackage{enumitem}
\usepackage{subcaption}
\usepackage{graphicx,subfig}
\usepackage{cleveref}
\usepackage{caption}
\usepackage{float}

\usepackage[utf8]{inputenc} %
\usepackage[T1]{fontenc}    %
\usepackage{hyperref}       %
\usepackage{url}            %
\usepackage{booktabs}       %
\usepackage{amsfonts}       %
\usepackage{nicefrac}       %
\usepackage{microtype}      %
\usepackage{graphicx}
\usepackage{subcaption} 
\usepackage{amssymb}
\usepackage{fdsymbol}
\usepackage{wrapfig}
\usepackage{lipsum}
\usepackage{enumitem}
\usepackage{stackengine}
\usepackage[font=small,labelfont=bf]{caption}
\usepackage{color}
\usepackage{adjustbox}

\usepackage{rotating}
\usepackage{makecell}

\definecolor{blanchedalmond}{rgb}{1.0, 0.92, 0.8}
\definecolor{carmine}{rgb}{0.59, 0.0, 0.09}
\definecolor{lightblue}{rgb}{0.22,0.45,0.70}%

\renewcommand{\mathbf}{\boldsymbol}

\makeatletter
\def\Ddots{\mathinner{\mkern1mu\raise\p@
\vbox{\kern7\p@\hbox{.}}\mkern2mu
\raise4\p@\hbox{.}\mkern2mu\raise7\p@\hbox{.}\mkern1mu}}
\makeatother

\definecolor{amaranth}{rgb}{0.9, 0.17, 0.31}
\definecolor{antiquebrass}{rgb}{0.8, 0.58, 0.46}
\definecolor{antiquefuchsia}{rgb}{0.57, 0.36, 0.51}
\definecolor{chromeyellow}{rgb}{0.31, 0.47, 0.26}

\newtcolorbox{AIbox}[2][]{aibox,title=#2,#1}
\definecolor{lightblue}{rgb}{0.22,0.45,0.70}%
\definecolor{Gray}{gray}{0.95}
\definecolor{Cornsilk}{rgb}{1.0, 0.97, 0.86}
\definecolor{posgreen}{RGB}{0,150,80}
\definecolor{negred}{RGB}{210,55,55}
\newcommand{\upval}[1]{\raisebox{-0.35ex}{\scriptsize\textcolor{posgreen}{+#1}}}
\newcommand{\downval}[1]{\raisebox{-0.35ex}{\scriptsize\textcolor{negred}{-#1}}}
\definecolor{posgreen}{RGB}{0,150,80}
\definecolor{negred}{RGB}{210,55,55}

\usepackage{amsmath}

\usepackage[all]{hypcap}

\title{%
  \makebox[\linewidth][l]{%
    \raisebox{-0.40\height}{%
      \includegraphics[
        height=1.45cm,
        keepaspectratio
      ]{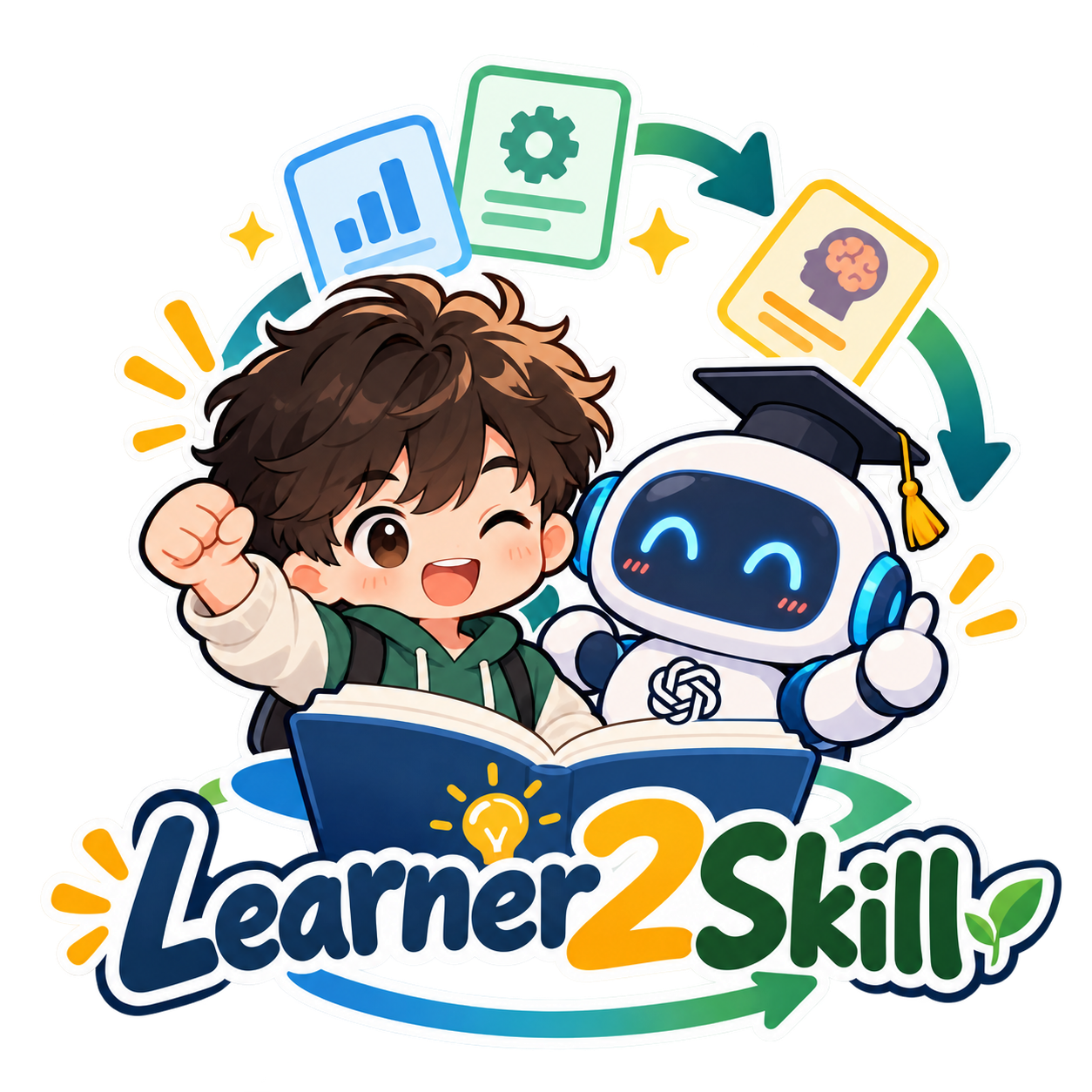}%
    }%
    \hspace{4mm}%
    \parbox[c]{0.84\linewidth}{%
      \raggedright
      \bfseries
      \fontsize{16}{19}\selectfont
      \mbox{From Learner Behavior to Reusable Skills}\\[1mm]
      \mbox{for Effective and Efficient Learner Simulation}%
    }%
  }%
}

\author[1]{Zijian Chen}
\author[2]{Zheng Zhang}
\author[1]{Miao Jia}
\author[1]{Xingchen Hu}
\author[3]{Weibo Gao\textsuperscript{\dagger}}
\author[4]{Linan Yue}

\affil[1]{National University of Defense Technology}
\affil[2]{Nanyang Technological University}
\affil[3]{\mbox{The Hong Kong Polytechnic University}}
\affil[4]{Southeast University}

\correspondingauthor{%
\textsuperscript{\dagger}Corresponding Author: Weibo Gao (\href{mailto:iamwebgao@gmail.com}{iamwebgao@gmail.com}).
\quad
{\small\textit{This work was completed in August 2026.}}
}

\begin{document}

\begin{abstract}
Learner simulation aims to reproduce how a particular learner behaves on new tasks. Although Large Language Models (LLMs) can generate increasingly fine-grained learning behaviors, existing approaches often need to repeatedly process a growing interaction history to reconstruct the learner. This introduces additional context and inference costs and makes the acquired learner-specific simulation capability difficult to reuse across different LLMs. We therefore propose \textbf{Learner2Skill}, which externalizes the simulation capability acquired from historical interactions into a persistent and reusable \textbf{Simulation Skill}. The Skill captures the learner's current learning state and recurring response patterns, evolves as new real interactions arrive, and can be adapted to a new LLM through lightweight executor calibration without reconstructing the learner from scratch. Experiments show that Learner2Skill more faithfully reproduces fine-grained learner behavior while reducing overall token cost, and that the same constructed Skills can be effectively reused across different LLM executors.
\end{abstract}

\schoollogos
\maketitle

\vspace{3mm}
\section{Introduction}

Learner simulation aims to reproduce how a particular learner behaves when facing new learning tasks~\citep{zhao2023simulating,scarlatos2026simulated}. A useful learner simulator should capture more than whether the learner ultimately answers correctly: it should also reflect whether the learner attempts the task, how the solution process unfolds, what errors or corrections appear along the way, and what answer is produced~\citep{asano2025can,koutcheme2026teaching}. Such fine-grained simulation provides a low-cost, controllable, and repeatable environment for developing and evaluating personalized teaching strategies without repeatedly involving real learners~\citep{markel2023gpteach,lu2024generative}. Recent advances in Large Language Models (LLMs) have made this increasingly feasible, as LLM-based agents can generate rich, multi-step responses conditioned on learner information~\citep{xu2024eduagent,liu2024personality}.

\begin{figure*}[t]
    \centering
    \includegraphics[width=1\textwidth]{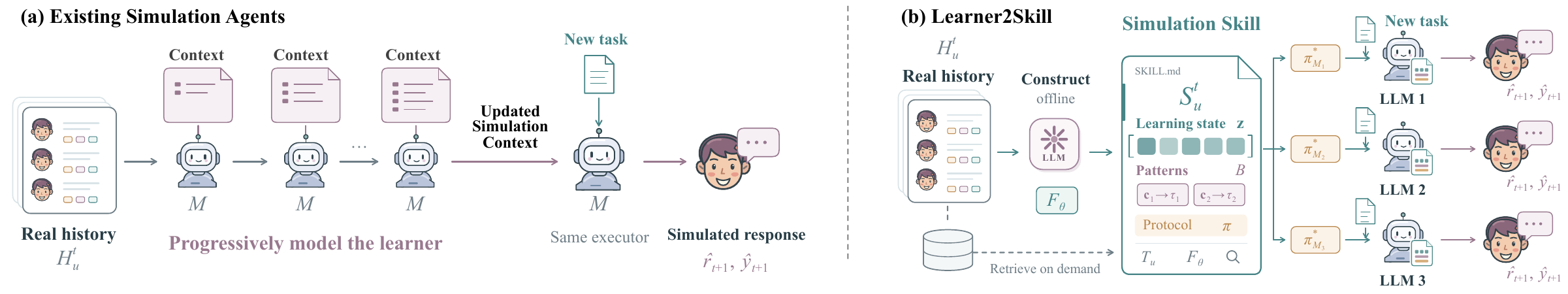}
    \caption{
Comparison of (a) existing LLM simulation agents that progressively model learners from interaction histories, and (b) Learner2Skill, which constructs reusable Simulation Skills for cross-executor simulation and online evolution.
}
    \label{fig:intro}
\end{figure*}

Existing LLM-based learner simulators, however, typically build their understanding of a learner by repeatedly processing the learner's interaction history, as illustrated in Figure~\ref{fig:intro}(a). As new interactions accumulate, learner profiles, memories, or learning states are progressively constructed and updated before being used for future simulation~\citep{gao2025agent4edu,duan2026history,xu2025classroom}. This process becomes increasingly expensive as the history grows. More importantly, the learner-specific simulation capability acquired from that history is usually coupled to the current agent and its underlying executor. The historical interactions themselves can be preserved, but when the LLM or executor changes, the same history may still need to be processed again to reconstruct an effective representation of the learner. In other words, existing approaches preserve the learner's experience, but provide no explicit mechanism for preserving and reusing the \emph{simulation capability acquired from that experience}.

This observation raises a natural question: \textit{if historical interactions have already revealed what a learner can currently do and how that learner typically responds, can this acquired capability be preserved and directly reused in future simulation?} We address this question with \textbf{Learner2Skill}, illustrated in Figure~\ref{fig:intro}(b), which organizes learner-specific simulation capability into a persistent and reusable \textbf{Simulation Skill}. Rather than repeatedly reconstructing the learner from the complete history, Learner2Skill distills the information most relevant to future simulation into the Skill, including the learner's current \textbf{learning state} and recurring \textbf{response patterns}, such as typical solution strategies, characteristic errors, corrections, and stopping tendencies. The complete real interaction history is retained separately as external evidence and retrieved only when finer-grained information is needed. Future simulations can therefore begin from an already constructed Skill while preserving access to the underlying historical evidence.

Learner2Skill supports this reusable representation through a complete lifecycle. \textbf{Skill Construction} first extracts the learner's current learning state and repeatedly supported response patterns from real historical interactions, while keeping cognitive modeling and historical retrieval available as on-demand resources. Because different LLMs may interpret and use the same learner information differently, \textbf{Executor Calibration} then uses a small amount of real interaction feedback from multiple learners to adapt a shared execution protocol, without changing the constructed learner information itself. When the underlying LLM changes, the existing Skills can therefore be retained and only lightweight calibration needs to be repeated. During deployment, \textbf{Skill Evolution} further updates the learner's state and response patterns as new real interactions arrive, allowing the Skill to remain aligned with the learner over time. This design explicitly separates learner-specific simulation capability from the executor that uses it: the Skill preserves who the learner is and how they tend to behave, while the executor determines how that information is used for the current task.

We evaluate Learner2Skill on \textbf{LearnerTrace-2K}, a temporally ordered dataset containing 52,566 mathematics and physics interactions from 2,000 learners, including learners' main problem-solving ideas and intermediate steps. Following the interaction chronology, each learner's trajectory is used sequentially for Skill Construction, Executor Calibration, and final evaluation. Under the same LLM executor, Learner2Skill achieves the best performance on five of six fine-grained simulation metrics while reducing overall token cost. The same constructed Skills can also be directly reused across different LLM executors without repeating Skill Construction. Ablation studies further show that learning state, response patterns, executor calibration, and online evolution contribute complementary benefits to different aspects of simulation. Finally, learner responses generated by Learner2Skill improve several computerized adaptive testing methods when used as additional training data. These results suggest that learner-specific simulation capability can be externalized from a particular LLM executor, preserved as a reusable Skill, and carried forward across subsequent learner simulation.

\section{Related Work}
\paragraph{\textbf{LLM-Based Human and Social Simulation}}
Large language models have increasingly been used to simulate human behavior at both individual and societal scales. Early studies investigate whether LLMs can reproduce human response distributions or support interactive societies populated by persona-driven agents~\citep{argyle2023outofone,aher2023simulatemultiple,park2023generativeagents}. Subsequent work scales such simulations to richer environments and larger populations~\citep{yang2024oasis,piao2025agentsociety}, while recent research increasingly grounds simulated agents in real human data. SocioVerse aligns agent societies with large-scale real-user populations~\citep{zhang2025socioverse}, whereas Generative Agent Simulations of 1,000 People and PersonaTwin construct person-specific agents from rich individual information to reproduce the behaviors of particular humans~\citep{park2024generative1000}. Recent work further moves beyond static personas toward dynamic behavioral simulation: Agentopia studies long-term agent development through years of simulated social experience, GRAPHIA aligns individual interactions and emergent social structures with real social graphs, and OdysSim develops behavior-oriented foundation models specifically for human simulation~\citep{wang2026agentopia,ji2026graphia,zhou2026odyssim}. These developments indicate a broader shift from plausible role-playing toward grounded, person-specific, and temporally evolving human simulation. Learner simulation represents a particularly demanding instance of this setting, because the target learner continuously changes through learning and must be reproduced not only at the level of high-level choices, but also in task-specific attempts, intermediate problem-solving behavior, characteristic errors, and learning outcomes.

\paragraph{\textbf{LLM-Based Learner Simulation}}
Early LLM-based learner simulation mainly used predefined personas, knowledge levels, or learner attributes to generate heterogeneous student behaviors for applications such as teacher training, item evaluation, and intelligent tutoring ~\citep{markel2023gpteach,lu2024generative,benedetto2024using,liu2024personality,nguyen2025qg,jin2025teachtune}. More recent studies construct learner agents from authentic learning data~\citep{xu2024eduagent,gao2025agent4edu}, covering classroom-level simulation~\citep{xu2025classroom,zhang2025simulating,gao2026theater}, tutoring dialogues~\citep{duan2026history,scarlatos2026simulated,zheng2025cognitive}, and domain-specific settings such as programming~\citep{zhan2025coderagent,duan2026kaser,koutcheme2026teaching}.
Most closely related to our work is \emph{individual learner response simulation}. Agent4Edu and Embracing Imperfection characterize individual learners from historical records and simulate their subsequent responses~\citep{gao2025agent4edu,wu2025embracing}, while SMART and One LLM Does Not Simulate All Students study the alignment between simulated behavior and learner ability~\citep{scarlatos2025smart,que2026one}. Empirical studies further show that LLM-generated solutions can differ systematically from real learner behavior even when final answers are correct~\citep{asano2025can}. Existing learner agents typically acquire learner-specific simulation capability through progressive processing of interaction histories, limiting both efficiency and reuse across executors. Learner2Skill instead preserves this capability as a persistent and reusable Simulation Skill.

\paragraph{\textbf{Skills for LLM Agents}}
Recent LLM-agent research increasingly studies how experience can be externalized into reusable knowledge, workflows, or skills. ExpeL, Agent Workflow Memory, and Voyager extract reusable guidance or executable procedures from previous trajectories~\citep{zhao2024expel,wang2024agent,wang2023voyager,yue2025don,wang2026ai,gong2026guided}, while SkillX, SAGE, SkillRL, and Skill1 further investigate construction, selection, transfer, and evolution of reusable skill libraries~\citep{wang2026reinforcement,wang2026skillx,xia2026skillrl,shi2026skill1}. Related work also explores improving externally represented skills from execution experience and transferring them across executors or environments~\citep{liu2026skillrevise}, as well as progressively internalizing external skills into model parameters~\citep{lu2026skill0}. These developments establish reusable skills as an emerging abstraction for preserving capabilities acquired from experience.
We apply this abstraction to \emph{learner-specific simulation capability}: rather than preserving how an agent solves a task, a Simulation Skill preserves how a particular learner tends to respond, allowing the acquired capability to be maintained and reused across LLM executors.

\section{The Proposed Learner2Skill Method}
\label{sec:method}


We focus on \textbf{individual learner response simulation}: given the previously observed real interactions of a particular learner, the goal is to simulate how the same learner would respond to a new learning task~\citep{duan2026history,scarlatos2026simulated}.
For learner $u$, we denote the real interaction history observed up to time step $t$ as
\begin{equation}
\mathcal H_u^t
=
\{(e_i,r_i,y_i)\}_{i=1}^{t},
\label{eq:history}
\end{equation}
where $e_i$ denotes a learning task, such as a mathematics problem, $r_i$ denotes the learner's complete response behavior on that task, and $y_i$ denotes the corresponding observable outcome. Specifically, we consider four simulation targets: 
(1) whether the learner attempts the task; 
(2) the response process, such as the solution strategy used, intermediate errors, corrections, or verification; 
(3) the final answer; and 
(4) the resulting performance, such as correctness or task score.
Given a new task $e_{t+1}$, the objective is therefore to generate the corresponding learner response and outcome $(\hat r_{t+1},\hat y_{t+1})$. Throughout simulation, only real interactions observed before the current task are available.

Existing LLM-based learner agents typically process a learner's interaction history step by step to construct and update learner-specific simulation context~\citep{gao2025agent4edu,duan2026history,xu2025classroom}. As the trajectory grows, this requires continued processing of historical interactions, while the acquired learner-specific simulation capability is typically coupled to the current agent and its executor, making it difficult to directly reuse across different executors.

In contrast, the proposed Learner2Skill explicitly organizes the learner-specific simulation capability acquired from historical interactions into a persistent and reusable \textbf{Simulation Skill}:
\begin{equation}
\mathcal S_u^t
=
\operatorname{Skill}
\left(
\mathcal H_u^t
\right).
\label{eq:skill_construction}
\end{equation}
Given a new task $e_{t+1}$, an executor $M$ uses the existing Skill to generate
\begin{equation}
(\hat r_{t+1},\hat y_{t+1})
\sim
q_{M,\pi}
\left(
r,y
\mid
e_{t+1},
\mathcal S_u^t
\right),
\label{eq:skill_simulation}
\end{equation}
where $\pi$ denotes the execution protocol through which the executor uses the Skill for learner simulation. A formal comparison between the conventional learner-agent pipeline and the Learner2Skill pipeline is provided in Appendix A.

\begin{figure*}[t]
    \centering
    \includegraphics[width=1\textwidth]{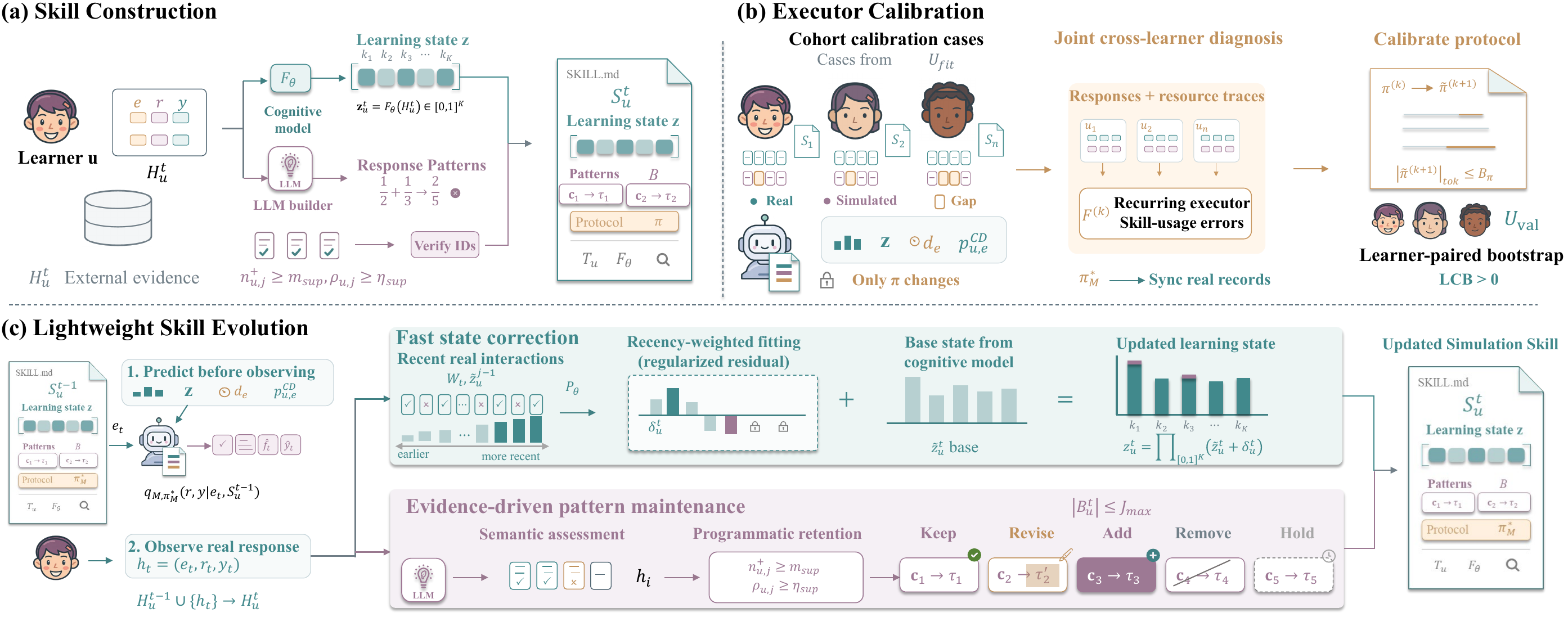}
    \caption{
    Overview of Learner2Skill. (a) Construct learner-specific Skills from real histories. (b) Calibrate a shared execution protocol through cross-learner diagnosis and held-out validation. (c) Update learner states and response patterns from new real interactions, with model parameters and the calibrated protocol fixed.
    }
    \label{fig:learner2skill_overview}
\end{figure*}

This formulation leads to three questions: 
\textit{How can a reusable Simulation Skill be constructed from historical learner interactions? 
How can an existing Skill be adapted to different downstream executors? 
And how can the Skill remain aligned with the learner as new real interactions arrive?}
Learner2Skill addresses these questions through \textbf{Skill Construction}, \textbf{Executor Calibration}, and \textbf{Skill Evolution}, respectively, as illustrated in Figure~\ref{fig:learner2skill_overview}.

\subsection{Constructing a Reusable Simulation Skill}
\label{sec:construction}

Skill Construction is performed offline before downstream learner simulation, independently of the downstream LLM executor that will later use the Skill for simulation. Given the observed real interaction history $\mathcal H_u^t$ of learner $u$, our goal is to extract learner-specific information that is useful for future response simulation and organize it into a reusable Simulation Skill. Specifically, we characterize two complementary aspects: \textbf{what the learner can currently do, represented by the learning state, and how the learner typically responds, represented by response patterns}.

Learning tasks typically involve one or more knowledge concepts, such as quadratic equations or function transformations in a mathematics problem~\citep{wang2023dynamic}. We therefore represent the learner's \textbf{learning state} through knowledge-concept mastery. Because such mastery is latent and cannot be directly observed, we use a pretrained cognitive model $F_\theta$ to estimate the learner's state from the real interactions observed up to time $t$~\citep{pandey2019self,liu2019exploiting}:
\begin{equation}
\mathbf z_u^t
=
F_\theta(\mathcal H_u^t)
\in[0,1]^K,
\label{eq:learning-state}
\end{equation}
where $K$ denotes the number of knowledge concepts, and the $k$-th dimension of $\mathbf z_u^t$ represents the learner's estimated mastery of concept $k$.

The cognitive model is also retained as a \textbf{tool} associated with the Skill for subsequent simulation. For the current task, the tool provides the relevant dimensions of the learner's learning state, the task difficulty, and the predicted probability that the learner will correctly complete the task~\citep{lord2012applications,scarlatos2025smart}. When additional cognitive information is needed, the executor can further query the cognitive model on demand. The construction, training, and use of the cognitive model are detailed in Appendix D.

The learning state captures the learner's current knowledge mastery, but similar learning states do not necessarily imply similar response behavior. Two learners with comparable knowledge states may still adopt different solution strategies, repeatedly make different types of errors, or exhibit different correction, verification, and stopping tendencies~\citep{liu2024personality,wu2025embracing,asano2025can}. We therefore use an offline LLM as the Skill builder to extract the learner's \textbf{response patterns} from the real interaction history~\citep{duan2026history}:
\begin{equation}
\mathcal B_u^t
=
\{(c_{u,j},\tau_{u,j})\}_{j=1}^{J_u^t},
\label{eq:response-patterns}
\end{equation}
where $c_{u,j}$ specifies the condition under which a pattern occurs, such as the relevant knowledge concepts, task difficulty, or other observable task characteristics, and $\tau_{u,j}$ describes the learner's typical response behavior under that condition. Such behavior may include using a particular solution strategy, making a characteristic error at a certain step, completing only part of a solution, checking or correcting a response, or stopping before completion.

The Skill builder used to construct these response patterns is separate from the downstream LLM executor that later performs learner simulation. A behavior is not retained as a response pattern based on a single occurrence; it enters $\mathcal B_u^t$ only when it is repeatedly supported by multiple real interactions, preventing isolated behavior from being treated as a persistent learner characteristic. The extraction, consolidation, and retention of response patterns are detailed in Appendix B.2. 

Because the learning state and response patterns summarize historical behavior, they may not capture every detail needed for a new task. We therefore retain the complete real interaction history $\mathcal H_u^t$ as external evidence in an interaction database and access it through retrieval rather than loading the full trajectory into every simulation context. When needed, the executor can retrieve the learner's most recent interactions, past interactions involving the current knowledge concepts, past interactions related to the current task content, and anonymized responses from other learners to the same task. The retrieval mechanism is detailed in Appendix B.5. 

Based on the above information, we represent the Simulation Skill of learner $u$ at time $t$ as
\begin{equation}
\mathcal S_u^t
=
\left(
\texttt{SKILL.md},
\mathbf z_u^t,
\mathcal B_u^t,
\mathcal T_u
\right),
\label{eq:simulation-skill}
\end{equation}
where $\mathbf z_u^t$ and $\mathcal B_u^t$ denote the learner's current learning state and response patterns, respectively, and $\mathcal T_u$ denotes the tool interfaces associated with the Skill, including the cognitive model $F_\theta$ and the retrieval tool for accessing the interaction database.
The markdown file \texttt{SKILL.md} serves as the unified entry point and operational specification of the Skill. It describes how the learner-specific information and associated tools are organized and contains a shared execution protocol $\pi$ that specifies how a downstream executor uses them during simulation. All learner Skills initially share the same base execution protocol, which is subsequently adapted to the current executor through Executor Calibration in Section~\ref{sec:calibration}. The organization of the Simulation Skill and the \texttt{SKILL.md} template are detailed in Appendix B.3. 

To keep the Skill compact as the interaction history grows, Learner2Skill maintains only the learner's \textbf{current learning state} and currently supported response patterns rather than their historical versions. The complete real interaction trajectory remains in the external database, while the number of maintained response patterns is bounded by a fixed pattern budget, preventing the Skill size from growing linearly with the interaction history.

\subsection{Calibrating a Skill to an Executor}
\label{sec:calibration}

After Skill Construction, the learner information and associated tools required for subsequent simulation have already been organized in the Simulation Skill. When a downstream LLM executor is equipped with a learner's Skill, it first reads the execution protocol in \texttt{SKILL.md} and, given the current task and available learner information, determines how to use the learner information already provided for the current task and whether additional resources, such as further cognitive-model queries or retrieval from the interaction database, should be invoked. Different LLMs or agent executors may vary in how they interpret the learning state and response patterns, when they invoke the associated tools, how they combine the available evidence, and how faithfully they reflect that evidence in the generated response. We therefore use a small amount of real interaction feedback from multiple learners to calibrate the shared execution protocol $\pi$ in \texttt{SKILL.md} to the current executor $M$, while keeping the executor model parameters fixed.
Specifically, we collect a small number of real interactions from multiple learners as the calibration set:
\begin{equation}
\mathcal D^{\mathrm{cal}}
=
\{(u_i,e_i,r_i,y_i)\}_{i=1}^{N_c}
\label{eq:calibration_set}
\end{equation}
where each interaction contains a task together with the corresponding real learner response and outcome. These interactions are observed after Skill Construction and before final testing. Rather than learning a separate usage strategy for each learner, calibration aggregates feedback across learners to identify recurring errors made by the current executor when using the Skill. Because these errors reflect shared problems in how the executor follows the common execution protocol, only a small number of real interactions from each learner is needed to refine the protocol, reducing the cost of adapting existing learner Skills to a new executor.

The learners participating in calibration are divided into a protocol-fitting set $\mathcal U_{\mathrm{fit}}$ and an independent validation set $\mathcal U_{\mathrm{val}}$, with no learner shared between them. The former is used to identify Skill-usage errors that recur across learners and to propose protocol edits, whereas the latter is used only to determine whether a candidate protocol yields consistent improvements on learners not involved in protocol editing. The calibration data and learner split are detailed in Appendix E.1. 

At calibration round $k$, the fixed executor $M$ uses the constructed Skills of learners in $\mathcal U_{\mathrm{fit}}$ together with the current execution protocol $\pi^{(k)}$ to simulate the corresponding tasks through the simulation interface described in Appendix B.6. 
Following the current protocol, the executor uses the learner information and current-task information already provided and, when needed, further queries the cognitive model or retrieves relevant historical evidence from the interaction database. The real learner response and outcome are revealed only after simulation and are used as feedback. By comparing the simulated response with the real response and examining how the executor uses the information and tools already available through the Skill, we identify recurring \textbf{Skill-usage errors} in how it interprets, requests, combines, or reflects learner evidence.

Because the execution protocol is shared across all learner Skills, we update it only from Skill-usage errors that recur across different learners rather than from isolated errors in individual interactions. When calibration interactions are processed in multiple mixed-learner batches, similar errors identified across batches are first consolidated before protocol editing; details are provided in Appendix E.2. 
Let $\mathcal F^{(k)}$ denote the cross-learner Skill-usage errors identified at round $k$. A candidate protocol is obtained by
\begin{equation}
\tilde{\pi}^{(k+1)}
=
\operatorname{Edit}
\left(
\pi^{(k)},
\mathcal F^{(k)}
\right),
\qquad
|\tilde{\pi}^{(k+1)}|_{\mathrm{tok}}
\le
B_{\pi}
\label{eq:protocol_edit}
\end{equation}
where $B_{\pi}$ constrains the protocol length so that the protocol remains compact across calibration rounds. The protocol-editing procedure is detailed in Appendix E.3. 

Each candidate protocol is then compared with the current protocol using calibration interactions from $\mathcal U_{\mathrm{val}}$. Validation computes paired improvements at the learner level and further applies a one-sided paired bootstrap to assess whether the improvement is consistent. A candidate is accepted only when the lower confidence bound of the paired improvement is greater than zero. Calibration terminates when no further candidate is accepted or the maximum number of calibration rounds $R_{\mathrm{cal}}$ is reached, yielding an executor-specific shared execution protocol $\pi_M^\star$. All learner Skills under executor $M$ then use this calibrated protocol. The validation procedure is detailed in Appendix E.4. 

During calibration, the learning state, response patterns, and accessible real historical evidence in the Skill remain unchanged; only the execution protocol is updated. After the final $\pi_M^\star$ is selected, the real interactions observed during calibration are incorporated into the corresponding learners' real interaction histories in chronological order and used to synchronize their Skills, so that the learning state and response patterns available at the beginning of final testing are consistent with the latest observed real interactions. The calibrated protocol $\pi_M^\star$ remains fixed during this synchronization. These calibration interactions are excluded from the final evaluation, and subsequent updates from newly observed real interactions follow the Skill Evolution procedure in Section~\ref{sec:evolution}. Details are provided in Appendix E.5. 

When the downstream LLM or agent executor changes from $M$ to a new executor $M'$, the constructed learner Skills and the real interaction histories stored in the database are retained~\citep{202609.0665}. The same low-cost calibration procedure is then applied to obtain a new execution protocol $\pi_{M'}^\star$ for the new executor. In this way, the learner-specific simulation capability extracted and organized in Stage~1 can be reused across executors without reconstructing the learner Skill from the complete interaction history ~\citep{liu2026skillrevise,wang2026skillx}. 
The theoretical analysis in Appendix G 
establishes when cross-executor Skill reuse gives Learner2Skill
a lower total token cost than existing LLM-based learner-simulation agents.

\subsection{Test-Time Skill Evolution with Real Interactions}
\label{sec:evolution}

After Skill Construction and Executor Calibration, Learner2Skill proceeds to online simulation. As new real interactions arrive, the learner's knowledge state and response behavior may continue to change, so the existing Skill needs to incorporate this new evidence over time. We therefore introduce lightweight \textbf{test-time Skill Evolution}, which progressively updates the learner's learning state and response patterns from newly observed real interactions~\citep{xia2026skillrl}.

For learner $u$ at time step $t$, simulation uses only the Skill available before the current real interaction:
\begin{equation}
(\hat r_t,\hat y_t)
\sim
q_{M,\pi_M^\star}
\left(
r,y
\mid
e_t,\mathcal S_u^{t-1}
\right).
\label{eq:online-prediction}
\end{equation}
After simulation, the real response $r_t$ and outcome $y_t$ are observed and appended to the learner's real interaction history $\mathcal H_u^t$. Skill Evolution uses only real interactions as update evidence; simulated responses are never treated as learner observations.

For the \textbf{learning state}, the frozen cognitive model $F_\theta$ first produces a new base estimate $\tilde{\mathbf z}_u^t$ from the updated real interaction history. To make the state responsive to recent learner behavior, we further estimate a learner-specific residual correction $\boldsymbol\delta_u^t$ from the most recent $L$ real interactions, yielding
\begin{equation}
\mathbf z_u^t
=
\Pi_{[0,1]^K}
\left(
\tilde{\mathbf z}_u^t
+
\boldsymbol\delta_u^t
\right).
\label{eq:learner-residual}
\end{equation}
The residual correction is fitted to recent real performance while being regularized in magnitude and temporal variation, allowing the learner state to adapt to recent evidence without updating the cognitive-model parameters. The recent-interaction objective and optimization procedure are detailed in Appendix F. 
For \textbf{response patterns}, we adopt an evidence-driven adaptive update mechanism. Each new real interaction is compared with the existing patterns to determine whether it provides additional support for an existing pattern, conflicts with one, or provides evidence for a new candidate pattern. Existing patterns are retained, revised, or removed according to accumulated evidence, while a new pattern is added only when it receives repeated support from multiple real interactions and satisfies the same retention criterion used during Skill Construction.

After each update, the latest $\mathbf z_u^t$ and $\mathcal B_u^t$ are stored in the Simulation Skill and used for subsequent learner simulation, while the complete real interaction history remains available in the external database for retrieval. Throughout Skill Evolution, the executor $M$, the calibrated execution protocol $\pi_M^\star$, and the underlying model parameters remain fixed.

\section{Experiments}
\label{sec:experiments}



{\textbf{\textit{Dataset.}}}
\label{sec:dataset}
Existing educational datasets primarily record learner outcomes or final answers,
while large-scale data with detailed problem-solving processes remain scarce.
We therefore collaborate with a real educational platform for high-school students
in China to construct \textbf{LearnerTrace-2K}, containing temporally ordered
mathematics and physics interactions with learners' main problem-solving ideas
and intermediate steps.
It contains \textbf{2,000 learners, 52,566 learner--exercise interactions,
2,076 exercises, and 709 knowledge concepts}.
For each learner, we preserve the original interaction order and chronologically
split the trajectory into {60\%/10\%/30\%} for
{Skill Construction}, {Calibration}, and  Evolution, respectively.
Dataset details are provided in Appendix H.1. 

{\textit{\textbf{Baselines}.}}
\label{sec:baselines}
We compare Learner2Skill with five LLM-based simulators: LLM-FullHistory, EduAgent~\citep{xu2024eduagent}, Agent4Edu~\citep{gao2025agent4edu}, CoderAgent~\citep{zhan2025coderagent}, and AAS~\citep{que2026one}; and four supervised predictors: KES~\citep{liu2019exploiting}, DKVMN~\citep{zhang2017dynamic}, SAKT~\citep{pandey2019self}, and DAISim~\citep{zhao2023simulating}. All LLM-based methods use the same executor and output format. Details are provided in Appendix H.2. 

{\textbf{\textit{Implementation Details.}}}
\label{sec:experimental_setup}
Executor Calibration uses the 10\% calibration split, with learners divided 70\%/30\% for protocol editing and validation. The cognitive model is trained only on the Skill Construction split and then fixed. RQ1 uses Claude Sonnet 4.6 for all LLM-based methods. RQ2 reuses the same constructed Skills across Claude Sonnet 4.6, Gemini 3.8 Flash, Qwen3.8 Flash, and DeepSeek V4 Flash, repeating only Executor Calibration. Semantic evaluation uses an independent GPT-5.5, and all experiments are repeated three times. Full settings are provided in Appendix H.3. 
Code is available at \url{https://github.com/WebGao/Learner2Skill-for-Effective-and-Efficient-Learner-Simulation}.

{\textbf{\textit{Evaluation Metrics.}}}
We evaluate \textbf{effectiveness} and \textbf{efficiency}.
For effectiveness, \textbf{Attempt}, \textbf{Process},
\textbf{Final-Answer}, and \textbf{Realized Correctness Fidelity} measure whether
the simulation reproduces the learner's attempt behavior, problem-solving process,
final answer, and realized success or failure, respectively.
\textbf{Outcome Prediction Accuracy} evaluates the explicit prediction of learner
performance, while \textbf{Prediction--Response Consistency} measures whether this
prediction agrees with the outcome realized by the generated response.
For efficiency, we report \textbf{Capability Acquisition Cost},
\textbf{Test-Time Cost}, their \textbf{Total Cost}, and
\textbf{Consistency Efficiency (CE)}, defined as Prediction--Response Consistency
per million generated tokens.
Full contents are provided in
Appendix H.4. 
\subsection{RQ1: Can We Simulate Learners More Effectively and Efficiently?}
\label{sec:rq1}

We first compare Learner2Skill with supervised learner-performance predictors
and LLM simulators on LearnerTrace-2K, examining both fine-grained
simulation effectiveness and generative cost.
For a fair comparison, all LLM-based methods use the same
\textbf{Claude Sonnet 4.6} simulation executor and are evaluated on the same
test interactions.

\begin{table*}[t]
\centering
\caption{
Main results on LearnerTrace-2K.
Effectiveness results are reported as mean $\pm$ standard deviation over three runs.
All LLM-based methods use \textbf{Claude Sonnet 4.6} as the simulation executor.
Higher is better for simulation effectiveness and cost-effectiveness, while lower is
better for simulation cost.
``--'' indicates that the metric is not applicable.
The best result is shown in \textbf{bold} and the second-best result is
\underline{underlined}.
}
\label{tab:main_results}

\setlength{\tabcolsep}{3.0pt}
\renewcommand{\arraystretch}{1.10}

\resizebox{\textwidth}{!}{
\begin{tabular}{lcccccccccc}
\toprule
& \multicolumn{6}{c}{\textbf{Simulation Effectiveness} ($\uparrow$)}
& \multicolumn{3}{c}{\textbf{Simulation Cost} ($\downarrow$)}
& \multicolumn{1}{c}{\textbf{Cost-Effectiveness} ($\uparrow$)} \\
\cmidrule(lr){2-7}
\cmidrule(lr){8-10}
\cmidrule(lr){11-11}

\textbf{Method}
& \textbf{Attempt}
& \textbf{Process}
& \textbf{Final Ans.}
& \textbf{Realized Corr.}
& \textbf{Outcome Acc.}
& \textbf{Consistency}
& \textbf{Acq.}
& \textbf{Test}
& \textbf{Total}
& \textbf{CE} \\

\midrule

\multicolumn{11}{l}{
\textit{\textbf{Supervised Learner-Performance Prediction}}
} \\

KES
& -- & -- & -- & --
& $52.92_{\pm 0.10}$
& -- & -- & -- & -- & -- \\

DKVMN
& -- & -- & -- & --
& $61.16_{\pm 0.10}$
& -- & -- & -- & -- & -- \\

SAKT
& -- & -- & -- & --
& $61.01_{\pm 0.09}$
& -- & -- & -- & -- & -- \\

DAISim
& -- & -- & -- & --
& $62.63_{\pm 0.05}$
& -- & -- & -- & -- & -- \\

\midrule

\multicolumn{11}{l}{
\textit{\textbf{LLM-Based Learner Simulation}}
} \\

LLM-FullHistory
& $\underline{99.22}_{\pm 0.07}$
& $50.66_{\pm 0.14}$
& $29.00_{\pm 0.08}$
& $46.88_{\pm 0.06}$
& $61.12_{\pm 0.13}$
& $60.48_{\pm 0.09}$
& \underline{227.170M}
& 125.585M
& \underline{352.755M}
& \underline{0.00171} \\

EduAgent
& $99.15_{\pm 0.09}$
& $52.16_{\pm 0.11}$
& $29.66_{\pm 0.09}$
& $47.33_{\pm 0.12}$
& $\underline{62.91}_{\pm 0.10}$
& $\underline{66.40}_{\pm 0.11}$
& 312.730M
& 133.770M
& 446.500M
& 0.00149 \\

Agent4Edu
& $98.13_{\pm 0.07}$
& $32.33_{\pm 0.13}$
& $37.10_{\pm 0.10}$
& $\mathbf{58.61}_{\pm 0.15}$
& $57.26_{\pm 0.09}$
& $55.26_{\pm 0.06}$
& 414.550M
& 118.455M
& 533.005M
& 0.00104 \\

CoderAgent
& $99.10_{\pm 0.13}$
& $\underline{53.77}_{\pm 0.13}$
& $33.24_{\pm 0.14}$
& $52.37_{\pm 0.15}$
& $59.25_{\pm 0.09}$
& $57.30_{\pm 0.09}$
& 351.234M
& 121.007M
& 472.241M
& 0.00121 \\

AAS
& $98.23_{\pm 0.08}$
& $53.70_{\pm 0.13}$
& $\underline{38.24}_{\pm 0.13}$
& $55.23_{\pm 0.12}$
& $60.12_{\pm 0.07}$
& $58.31_{\pm 0.10}$
& 403.277M
& \underline{117.532M}
& 520.809M
& 0.00112 \\

\midrule


\textbf{Learner2Skill}
& $\mathbf{99.64}_{\pm 0.06}$
& $\mathbf{54.19}_{\pm 0.10}$
& $\mathbf{40.13}_{\pm 0.13}$
& $\underline{57.71}_{\pm 0.07}$
& $\mathbf{63.43}_{\pm 0.09}$
& $\mathbf{71.54}_{\pm 0.11}$
& \textbf{182.590M}
& \textbf{116.570M}
& \textbf{299.160M}
& \textbf{0.00239} \\

\bottomrule
\end{tabular}
}

\end{table*}

\paragraph{\textbf{Simulation Effectiveness.}}
As shown in Table~\ref{tab:main_results}, the base LLM already exhibits meaningful
learner-simulation capability: LLM-FullHistory, which directly uses the complete
learner history, reaches 50.66 in Process Fidelity and 61.12 in Outcome Prediction
Accuracy, but remains limited in Final-Answer Fidelity (29.00) and
Prediction--Response Consistency (60.48).
While different learner-modeling methods improve particular behavioral dimensions,
Learner2Skill provides more consistent gains, achieving the best results in Attempt,
Process, Final-Answer, Outcome Prediction, and Prediction--Response Consistency.
Compared with LLM-FullHistory, it improves Final-Answer Fidelity from 29.00 to
40.13 and Prediction--Response Consistency from 60.48 to 71.54; its Outcome
Prediction Accuracy (63.43) also exceeds the strongest supervised predictor,
DAISim (62.63).
These results show that the Simulation Skill preserves strong outcome prediction
while more faithfully reproducing the learner's problem-solving process, final
answer, and the consistency between predicted and generated behavior.

\paragraph{\textbf{Simulation Cost and Cost-Effectiveness.}}
Learner2Skill improves simulation effectiveness while also reducing generative
cost, achieving the lowest Capability Acquisition Cost (182.590M), Test-Time Cost
(116.570M), and Total Cost (299.160M) among all LLM-based methods.
Its total cost is approximately \textbf{15.2\%} lower than the next-lowest method,
LLM-FullHistory (352.755M).
This efficiency gain reflects the reusable nature of the Simulation Skill:
although LLM-FullHistory repeatedly processes the complete learner history during
simulation, Learner2Skill acquires the learner-specific simulation capability once
and subsequently reuses the resulting Skill across interactions.
Combining effectiveness and efficiency, Learner2Skill achieves the highest CE of
\textbf{0.00239}, approximately \textbf{39.8\%} above the second-best result.
Together, these results show that organizing learner-specific simulation capability
as a reusable Simulation Skill improves fine-grained behavioral simulation while
reducing the cost of acquiring and repeatedly using that capability.

\subsection{RQ2: Can Simulation Skills Be Reused Across Different LLM Executors?}
\label{sec:rq2}

We next examine whether a Simulation Skill constructed once can be reused across
different LLM executors.
We first run the complete pipeline with \textbf{Claude Sonnet 4.6} to construct
the initial Simulation Skills, and then directly reuse these Skills with
\textbf{Gemini 3.8 Flash}, \textbf{Qwen3.8 Flash}, and
\textbf{DeepSeek V4 Flash}.
For each reuse setting, Skill Construction is not repeated; only
executor-specific calibration is performed before final testing.

\begin{table*}[t]
\centering
\caption{
Cross-executor reuse of Simulation Skills on LearnerTrace-2K.
\textbf{Initial} denotes the complete pipeline used to construct the Skills,
whereas \textbf{Reuse} directly applies the constructed Skills to a new executor
without repeating Skill Construction.
}
\label{tab:cross_executor}

\setlength{\tabcolsep}{3.4pt}
\renewcommand{\arraystretch}{1.10}

\resizebox{\textwidth}{!}{
\begin{tabular}{llcccccccccc}
\toprule
\multicolumn{2}{c}{\textbf{Configuration}}
& \multicolumn{6}{c}{\textbf{Simulation Fidelity} ($\uparrow$)}
& \multicolumn{4}{c}{\textbf{Simulation Cost} ($\downarrow$)} \\
\cmidrule(lr){1-2}
\cmidrule(lr){3-8}
\cmidrule(lr){9-12}

\textbf{Setting}
& \textbf{Executor}
& \textbf{Attempt}
& \textbf{Process}
& \textbf{Final Ans.}
& \textbf{Realized Corr.}
& \textbf{Outcome Acc.}
& \textbf{Consistency}
& \textbf{Build}
& \textbf{Calib.}
& \textbf{Test}
& \textbf{Total} \\
\midrule

\textbf{Initial}
& Claude Sonnet 4.6
& 99.64
& \underline{54.19}
& 40.13
& \textbf{57.71}
& \underline{63.43}
& 71.54
& 125.36M
& \underline{57.23M}
& 116.57M
& 299.16M \\

\midrule

\textit{Reuse}
& Gemini 3.8 Flash
& \underline{99.78}
& \textbf{54.53}
& \textbf{43.74}
& 57.22
& \textbf{63.73}
& \underline{75.21}
& 0
& 60.55M
& 132.13M
& 192.68M \\

\textit{Reuse}
& Qwen3.8 Flash
& 98.27
& 50.15
& 40.07
& 55.24
& 59.77
& 72.21
& 0
& \textbf{53.21M}
& \underline{103.12M}
& \underline{156.33M} \\

\textit{Reuse}
& DeepSeek V4 Flash
& \textbf{99.94}
& 48.93
& \underline{43.11}
& \underline{57.42}
& 60.86
& \textbf{75.22}
& 0
& 60.17M
& \textbf{80.60M}
& \textbf{140.77M} \\

\bottomrule
\end{tabular}
}

\begin{minipage}{0.99\textwidth}
\footnotesize
\end{minipage}
\end{table*}

As shown in Table~\ref{tab:cross_executor}, the same Simulation Skills remain
effective across all tested executors.
Gemini 3.8 Flash achieves the highest Process Fidelity (54.53),
Final-Answer Fidelity (43.74), and Outcome Prediction Accuracy (63.73), while
DeepSeek V4 Flash achieves the highest Prediction--Response Consistency (75.22).
Importantly, changing the executor does not require repeating the
125.36M-token Skill Construction stage.
Instead, the previously constructed Skills are retained and only
executor-specific calibration is performed.
The resulting total costs are 192.68M, 156.33M, and 140.77M tokens for
Gemini, Qwen, and DeepSeek, respectively, compared with 299.16M tokens for the
initial pipeline.
These results show that learner-specific simulation capability can be retained
across different executors, allowing a new LLM to reuse the existing Simulation
Skill rather than reconstructing it from the learner's history.

\enlargethispage{2\baselineskip}
\subsection{RQ3: What Makes Learner2Skill Effective?}
\label{sec:ablation}
\begin{table*}[t]
\centering
\caption{
Ablation study of Learner2Skill on LearnerTrace-2K.
Colored numbers indicate changes relative to the full model.
}
\label{tab:ablation}

\setlength{\tabcolsep}{6.0pt}
\renewcommand{\arraystretch}{1.12}

\resizebox{\textwidth}{!}{
\begin{tabular}{lcccccc}
\toprule
\textbf{Method}
& \textbf{Attempt} $\uparrow$
& \textbf{Process} $\uparrow$
& \textbf{Final Ans.} $\uparrow$
& \textbf{Realized Corr.} $\uparrow$
& \textbf{Outcome Acc.} $\uparrow$
& \textbf{Consistency} $\uparrow$ \\
\midrule

\textbf{Full Learner2Skill}
& \textbf{99.64}
& \textbf{54.19}
& \textbf{40.13}
& \textbf{57.71}
& \textbf{63.43}
& \textbf{71.54} \\

\midrule

w/o Learning State
& 99.13\,\downval{0.51}
& 56.80\,\upval{2.61}
& 40.03\,\downval{0.10}
& 53.80\,\downval{3.91}
& 62.03\,\downval{1.40}
& 70.51\,\downval{1.03} \\

w/o Response Patterns
& 99.98\,\upval{0.34}
& 35.05\,\downval{19.14}
& 39.64\,\downval{0.49}
& 58.50\,\upval{0.79}
& 62.60\,\downval{0.83}
& 70.18\,\downval{1.36} \\

w/o Executor Calibration
& 99.63\,\downval{0.01}
& 53.46\,\downval{0.73}
& 39.75\,\downval{0.38}
& 57.21\,\downval{0.50}
& 62.57\,\downval{0.86}
& 71.40\,\downval{0.14} \\

w/o Skill Evolution
& 99.53\,\downval{0.11}
& 52.83\,\downval{1.36}
& 39.38\,\downval{0.75}
& 56.86\,\downval{0.85}
& 62.51\,\downval{0.92}
& 71.37\,\downval{0.17} \\

\bottomrule
\end{tabular}
}

\end{table*}

We ablate the four key components of Learner2Skill: Learning State and Response Patterns in Skill Construction, Executor Calibration, and Skill Evolution. As shown in Table~\ref{tab:ablation}, Response Patterns are particularly important for Process Fidelity, while Learning State mainly improves outcome-related metrics. This is consistent with their roles: Response Patterns characterize how a learner typically behaves during problem solving, whereas Learning State reflects the learner's current capability. Executor Calibration and Skill Evolution provide further consistent gains by improving Skill use and keeping learner-specific information aligned with newly observed behavior. Together, these results confirm the complementary roles of the four components.


\subsection{RQ4: Can Learner2Skill Benefit Real Educational Applications?}
\label{sec:rq4}



We evaluate whether learner responses generated by Learner2Skill can improve
Computerized Adaptive Testing (CAT), a practical application that estimates learner
proficiency using a limited number of adaptively selected exercises. We consider
three CAT strategies: FSI~\citep{lord2012applications},
KLI~\citep{chang1996global}, and MAAT~\citep{bi2020quality},
with test lengths of 5 and 10 exercises.

\begin{wrapfigure}[10]{r}{0.54\columnwidth}
    \vspace{-5pt}
    \centering
    \includegraphics[width=\linewidth]{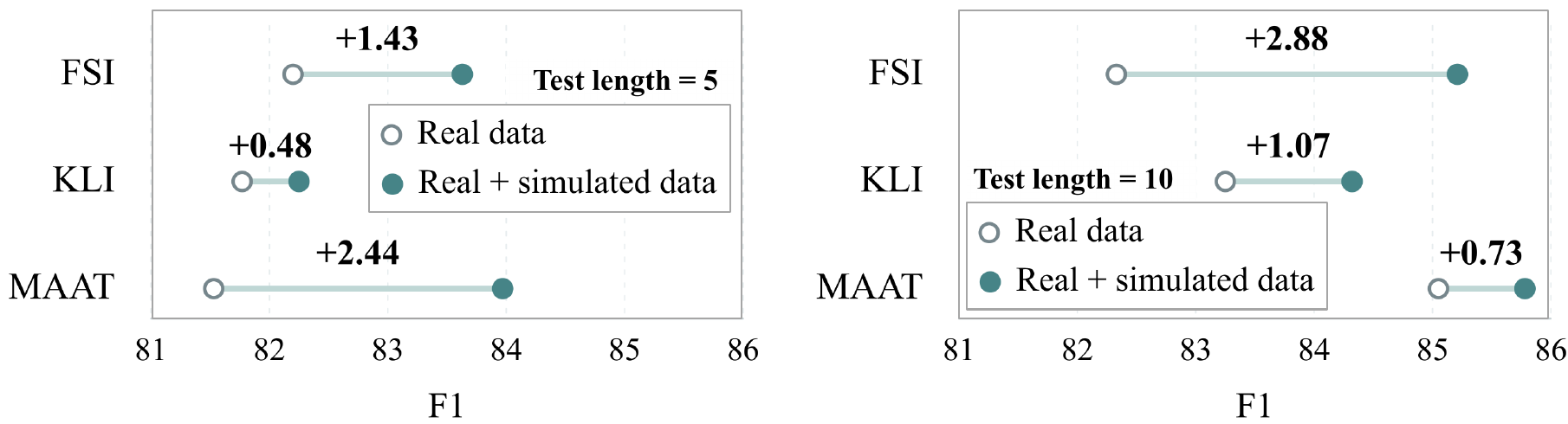}
    \captionsetup{
        justification=raggedright,
        singlelinecheck=false,
        font=small,
        skip=4pt
    }
    \caption{
        CAT performance with generated data.
        Numbers above the lines denote F1 improvements after augmentation.
    }
    \label{fig:cat_application}
    \vspace{-5pt}
\end{wrapfigure}

\begin{sloppypar}
For each strategy, we compare training on the original real learner data with
training on the same data augmented by simulated learner responses. The CAT
strategy remains unchanged, and both settings are evaluated on the same held-out
real learners.
Figure~\ref{fig:cat_application} shows higher F1 scores with simulated-data
augmentation across all three strategies and both test lengths, with gains ranging
from 0.48 to 2.88. These results suggest that Learner2Skill-generated responses
provide useful additional supervision for CAT, improving performance on real
learners without changing the adaptive testing strategy.
\end{sloppypar}
\section{Conclusion}

We introduced \textbf{Learner2Skill}, which organizes learner-specific simulation capability acquired from historical interactions into persistent and reusable \textbf{Simulation Skills}, avoiding repeated reconstruction of the learner in subsequent simulation. Through Skill Construction, Executor Calibration, and Skill Evolution, Learner2Skill supports the construction, cross-executor reuse, and continual updating of learner-specific information. Experiments show that it more faithfully reproduces fine-grained learner behavior while reducing overall token cost, and that constructed Skills can be directly reused across different LLM executors. The generated learner responses also provide useful supervision for practical educational applications. Overall, Learner2Skill provides a way to decouple learner-specific simulation capability from a particular LLM and preserve it for continued reuse.

\clearpage
\bibliography{main}

@inproceedings{asano2025can,
  title={Can LLMs simulate the same correct solutions to free-response math problems as real students?},
  author={Asano, Yuya and Litman, Diane and Walker, Erin},
  booktitle={Proceedings of the 2025 Conference on Empirical Methods in Natural Language Processing},
  pages={16347--16376},
  year={2025}
}

@inproceedings{benedetto2024using,
  title={Using LLMs to simulate students’ responses to exam questions},
  author={Benedetto, Luca and Aradelli, Giovanni and Donvito, Antonia and Lucchetti, Alberto and Cappelli, Andrea and Buttery, Paula},
  booktitle={Findings of the Association for Computational Linguistics: EMNLP 2024},
  pages={11351--11368},
  year={2024}
}

@inproceedings{duan2026kaser,
  title={Kaser: Knowledge-aligned student error simulator for open-ended coding tasks},
  author={Duan, Zhangqi and Fernandez, Nigel and Lan, Andrew},
  booktitle={Proceedings of the 64th Annual Meeting of the Association for Computational Linguistics (Volume 1: Long Papers)},
  pages={39988--40006},
  year={2026}
}

@article{duan2026history,
  title={Who Am I? History-Aware Profiles for Student Simulation in Tutoring Dialogues},
  author={Duan, Zhangqi and Huang, Shuyan and Scarlatos, Alexander and Lee, Jaewook and Woodhead, Simon and Lan, Andrew},
  journal={arXiv preprint arXiv:2605.30051},
  year={2026}
}

@inproceedings{gao2025agent4edu,
  title={Agent4edu: Generating learner response data by generative agents for intelligent education systems},
  author={Gao, Weibo and Liu, Qi and Yue, Linan and Yao, Fangzhou and Lv, Rui and Zhang, Zheng and Wang, Hao and Huang, Zhenya},
  booktitle={Proceedings of the AAAI Conference on Artificial Intelligence},
  volume={39},
  number={22},
  pages={23923--23932},
  year={2025}
}

@article{gao2026theater,
  title={Edu-Theater: A Data-Efficient Agent Framework for Scalable Learner Behavior Simulation through Staging Roll-Call},
  author={Gao, Weibo and Liu, Qi and Yue, Linan and Zhang, Zheng and Du, Yichao and Yao, Fangzhou and Yu, Ao and Huang, Zhenya and Wang, Shijin},
  journal={arXiv preprint arXiv:2606.15225},
  year={2026}
}

@inproceedings{jin2025teachtune,
  title={Teachtune: Reviewing pedagogical agents against diverse student profiles with simulated students},
  author={Jin, Hyoungwook and Yoo, Minju and Park, Jeongeon and Lee, Yokyung and Wang, Xu and Kim, Juho},
  booktitle={Proceedings of the 2025 CHI Conference on Human Factors in Computing Systems},
  pages={1--28},
  year={2025}
}

@article{koutcheme2026teaching,
  title={Teaching language models how to code like learners: Conversational serialization for student simulation},
  author={Koutcheme, Charles and Leinonen, Juho and Hellas, Arto},
  journal={arXiv preprint arXiv:2604.10720},
  year={2026}
}

@article{liu2026skillrevise,
  title={SkillRevise: Improving LLM-Authored Agent Skills via Trace-Conditioned Skill Revision},
  author={Liu, Yuxuan and Su, Zhaochen and Xie, Lingyun and Zhang, Yuhao and Zong, Qing and Guo, Jiahe and Xie, Zhongwei and Ji, Yiyan and Yim, Yauwai and Luo, Hongyu and others},
  journal={arXiv preprint arXiv:2606.01139},
  year={2026}
}

@inproceedings{liu2024personality,
  title={Personality-aware student simulation for conversational intelligent tutoring systems},
  author={Liu, Zhengyuan and Yin, Stella Xin and Lin, Geyu and Chen, Nancy},
  booktitle={Proceedings of the 2024 Conference on Empirical Methods in Natural Language Processing},
  pages={626--642},
  year={2024}
}

@inproceedings{lu2024generative,
  title={Generative students: Using llm-simulated student profiles to support question item evaluation},
  author={Lu, Xinyi and Wang, Xu},
  booktitle={Proceedings of the Eleventh ACM conference on learning@ Scale},
  pages={16--27},
  year={2024}
}

@article{lu2026skill0,
  title={Skill0: In-context agentic reinforcement learning for skill internalization},
  author={Lu, Zhengxi and Yao, Zhiyuan and Wu, Jinyang and Han, Chengcheng and Gu, Qi and Cai, Xunliang and Lu, Weiming and Xiao, Jun and Zhuang, Yueting and Shen, Yongliang},
  journal={arXiv preprint arXiv:2604.02268},
  year={2026}
}

@inproceedings{markel2023gpteach,
  title={Gpteach: Interactive ta training with gpt-based students},
  author={Markel, Julia M and Opferman, Steven G and Landay, James A and Piech, Chris},
  booktitle={Proceedings of the tenth acm conference on learning@ scale},
  pages={226--236},
  year={2023}
}

@inproceedings{nguyen2025qg,
  title={Qg-sms: enhancing test item analysis via student modeling and simulation},
  author={Nguyen, Bang and Du, Tingting and Yu, Mengxia and Angrave, Lawrence and Jiang, Meng},
  booktitle={Proceedings of the 63rd Annual Meeting of the Association for Computational Linguistics (Volume 1: Long Papers)},
  pages={26152--26168},
  year={2025}
}

@inproceedings{que2026one,
  title={One LLM Does Not Simulate All Students: Ability-Aware Student Simulation via Cognitive Diagnosis Guided LLM Assignment},
  author={Que, Huixing and Liu, Qi and Gao, Weibo and Huang, Zhenya},
  booktitle={Findings of the Association for Computational Linguistics: ACL 2026},
  pages={6067--6084},
  year={2026}
}

@inproceedings{scarlatos2025smart,
  title={Smart: Simulated students aligned with item response theory for question difficulty prediction},
  author={Scarlatos, Alexander and Fernandez, Nigel and Ormerod, Christopher and Lottridge, Susan and Lan, Andrew},
  booktitle={Proceedings of the 2025 Conference on Empirical Methods in Natural Language Processing},
  pages={25082--25105},
  year={2025}
}

@inproceedings{scarlatos2026simulated,
  title={Simulated Students in Tutoring Dialogues: Substance or Illusion?},
  author={Scarlatos, Alexander and Lee, Jaewook and Woodhead, Simon and Lan, Andrew},
  booktitle={Proceedings of the 64th Annual Meeting of the Association for Computational Linguistics (Volume 1: Long Papers)},
  pages={42349--42385},
  year={2026}
}

@article{shi2026skill1,
  title={Skill1: Unified evolution of skill-augmented agents via reinforcement learning},
  author={Shi, Yaorui and Chen, Yuxin and Lu, Zhengxi and Miao, Yuchun and Liu, Shugui and Gu, Qi and Cai, Xunliang and Wang, Xiang and Zhang, An},
  journal={arXiv preprint arXiv:2605.06130},
  year={2026}
}

@article{wang2026skillx,
  title={Skillx: Automatically constructing skill knowledge bases for agents},
  author={Wang, Chenxi and Yu, Zhuoyun and Xie, Xin and Yao, Wuguannan and Fang, Runnan and Qiao, Shuofei and Cao, Kexin and Zheng, Guozhou and Qi, Xiang and Zhang, Peng and others},
  journal={arXiv preprint arXiv:2604.04804},
  year={2026}
}

@article{wang2023voyager,
  title={Voyager: An open-ended embodied agent with large language models},
  author={Wang, Guanzhi and Xie, Yuqi and Jiang, Yunfan and Mandlekar, Ajay and Xiao, Chaowei and Zhu, Yuke and Fan, Linxi and Anandkumar, Anima},
  journal={arXiv preprint arXiv:2305.16291},
  year={2023}
}

@inproceedings{wang2026reinforcement,
  title={Reinforcement learning for self-improving agent with skill library},
  author={Wang, Jiongxiao and Yan, Qiaojing and Wang, Yawei and Tian, Yijun and Mishra, Soumya Smruti and Xu, Zhichao and Gandhi, Megha and Xu, Panpan and Cheong, Lin Lee},
  booktitle={Proceedings of the 64th Annual Meeting of the Association for Computational Linguistics (Volume 1: Long Papers)},
  pages={1529--1550},
  year={2026}
}

@article{wang2024agent,
  title={Agent workflow memory},
  author={Wang, Zora Zhiruo and Mao, Jiayuan and Fried, Daniel and Neubig, Graham},
  journal={arXiv preprint arXiv:2409.07429},
  year={2024}
}

@inproceedings{wu2025embracing,
  title={Embracing imperfection: Simulating students with diverse cognitive levels using LLM-based agents},
  author={Wu, Tao and Chen, Jingyuan and Lin, Wang and Li, Mengze and Zhu, Yumeng and Li, Ang and Kuang, Kun and Wu, Fei},
  booktitle={Proceedings of the 63rd Annual Meeting of the Association for Computational Linguistics (Volume 1: Long Papers)},
  pages={9887--9908},
  year={2025}
}

@article{xia2026skillrl,
  title={Skillrl: Evolving agents via recursive skill-augmented reinforcement learning},
  author={Xia, Peng and Chen, Jianwen and Wang, Hanyang and Liu, Jiaqi and Zeng, Kaide and Wang, Yu and Han, Siwei and Zhou, Yiyang and Zhao, Xujiang and Chen, Haifeng and others},
  journal={arXiv preprint arXiv:2602.08234},
  year={2026}
}

@article{xu2024eduagent,
  title={Eduagent: Generative student agents in learning},
  author={Xu, Songlin and Zhang, Xinyu and Qin, Lianhui},
  journal={arXiv preprint arXiv:2404.07963},
  year={2024}
}

@inproceedings{xu2025classroom,
  title={Classroom simulacra: Building contextual student generative agents in online education for learning behavioral simulation},
  author={Xu, Songlin and Wen, Hao-Ning and Pan, Hongyi and Dominguez, Dallas and Hu, Dongyin and Zhang, Xinyu},
  booktitle={Proceedings of the 2025 CHI Conference on Human Factors in Computing Systems},
  pages={1--26},
  year={2025}
}

@article{zhan2025coderagent,
  title={Coderagent: Simulating student behavior for personalized programming learning with large language models},
  author={Zhan, Yi and Liu, Qi and Gao, Weibo and Zhang, Zheng and Wang, Tianfu and Shen, Shuanghong and Lu, Junyu and Huang, Zhenya},
  journal={arXiv preprint arXiv:2505.20642},
  year={2025}
}

@inproceedings{zhang2025simulating,
  title={Simulating classroom education with llm-empowered agents},
  author={Zhang, Zheyuan and Zhang-Li, Daniel and Yu, Jifan and Gong, Linlu and Zhou, Jinchang and Hao, Zhanxin and Jiang, Jianxiao and Cao, Jie and Liu, Huiqin and Liu, Zhiyuan and others},
  booktitle={Proceedings of the 2025 conference of the nations of the americas chapter of the association for computational linguistics: Human language technologies (volume 1: Long papers)},
  pages={10364--10379},
  year={2025}
}

@inproceedings{zhao2024expel,
  title={Expel: Llm agents are experiential learners},
  author={Zhao, Andrew and Huang, Daniel and Xu, Quentin and Lin, Matthieu and Liu, Yong-Jin and Huang, Gao},
  booktitle={Proceedings of the AAAI Conference on Artificial Intelligence},
  volume={38},
  number={17},
  pages={19632--19642},
  year={2024}
}

@article{zheng2025cognitive,
  title={Cognitive Echo: Enhancing think-aloud protocols with LLM-based simulated students},
  author={Zheng, Longwei and He, Anna and Qi, Changyong and Zhang, Haomin and Gu, Xiaoqing},
  journal={British Journal of Educational Technology},
  volume={56},
  number={5},
  pages={2019--2042},
  year={2025},
  publisher={Wiley Online Library}
}

@inproceedings{liu2019exploiting,
  title={Exploiting cognitive structure for adaptive learning},
  author={Liu, Qi and Tong, Shiwei and Liu, Chuanren and Zhao, Hongke and Chen, Enhong and Ma, Haiping and Wang, Shijin},
  booktitle={Proceedings of the 25th ACM SIGKDD international conference on knowledge discovery \& data mining},
  pages={627--635},
  year={2019}
}

@inproceedings{zhang2017dynamic,
  title={Dynamic key-value memory networks for knowledge tracing},
  author={Zhang, Jiani and Shi, Xingjian and King, Irwin and Yeung, Dit-Yan},
  booktitle={Proceedings of the 26th international conference on World Wide Web},
  pages={765--774},
  year={2017}
}

@article{pandey2019self,
  title={A self-attentive model for knowledge tracing},
  author={Pandey, Shalini and Karypis, George},
  journal={arXiv preprint arXiv:1907.06837},
  year={2019}
}

@inproceedings{zhao2023simulating,
  title={Simulating student interactions with two-stage imitation learning for intelligent educational systems},
  author={Zhao, Guanhao and Huang, Zhenya and Zhuang, Yan and Liu, Jiayu and Liu, Qi and Liu, Zhiding and Wu, Jinze and Chen, Enhong},
  booktitle={Proceedings of the 32nd ACM International Conference on Information and Knowledge Management},
  pages={3423--3432},
  year={2023}
}

@book{lord2012applications,
  title={Applications of item response theory to practical testing problems},
  author={Lord, Frederic M},
  year={2012},
  publisher={Routledge}
}

@article{chang1996global,
  title={A global information approach to computerized adaptive testing},
  author={Chang, Hua-Hua and Ying, Zhiliang},
  journal={Applied Psychological Measurement},
  volume={20},
  number={3},
  pages={213--229},
  year={1996},
  publisher={Sage Publications Sage CA: Thousand Oaks, CA}
}

@inproceedings{bi2020quality,
  title={Quality meets diversity: A model-agnostic framework for computerized adaptive testing},
  author={Bi, Haoyang and Ma, Haiping and Huang, Zhenya and Yin, Yu and Liu, Qi and Chen, Enhong and Su, Yu and Wang, Shijin},
  booktitle={2020 IEEE International Conference on Data Mining (ICDM)},
  pages={42--51},
  year={2020},
  organization={IEEE}
}

@article{wang2023dynamic,
  title={Dynamic cognitive diagnosis: An educational priors-enhanced deep knowledge tracing perspective},
  author={Wang, Fei and Huang, Zhenya and Liu, Qi and Chen, Enhong and Yin, Yu and Ma, Jianhui and Wang, Shijin},
  journal={IEEE Transactions on Learning Technologies},
  volume={16},
  number={3},
  pages={306--323},
  year={2023},
  publisher={IEEE}
}

@article{yue2025don,
  title={Don't Overthink It: A Survey of Efficient R1-style Large Reasoning Models},
  author={Yue, Linan and Du, Yichao and Wang, Yizhi and Gao, Weibo and Yao, Fangzhou and Wang, Li and Liu, Ye and Xu, Ziyu and Liu, Qi and Di, Shimin and others},
  journal={arXiv preprint arXiv:2508.02120},
  year={2025}
}

@article{argyle2023outofone,
  title={Out of one, many: Using language models to simulate human samples},
  author={Argyle, Lisa P and Busby, Ethan C and Fulda, Nancy and Gubler, Joshua R and Rytting, Christopher and Wingate, David},
  journal={Political Analysis},
  volume={31},
  number={3},
  pages={337--351},
  year={2023},
  publisher={Cambridge University Press}
}

@inproceedings{aher2023simulatemultiple,
  title={Using large language models to simulate multiple humans and replicate human subject studies},
  author={Aher, Gati V and Arriaga, Rosa I and Kalai, Adam Tauman},
  booktitle={International conference on machine learning},
  pages={337--371},
  year={2023},
  organization={PMLR}
}

@inproceedings{park2023generativeagents,
  title={Generative agents: Interactive simulacra of human behavior},
  author={Park, Joon Sung and O'Brien, Joseph and Cai, Carrie Jun and Morris, Meredith Ringel and Liang, Percy and Bernstein, Michael S},
  booktitle={Proceedings of the 36th annual acm symposium on user interface software and technology},
  pages={1--22},
  year={2023}
}

@article{yang2024oasis,
  title={Oasis: Open agent social interaction simulations with one million agents},
  author={Yang, Ziyi and Zhang, Zaibin and Zheng, Zirui and Jiang, Yuxian and Gan, Ziyue and Wang, Zhiyu and Ling, Zijian and Chen, Jinsong and Ma, Martz and Dong, Bowen and others},
  journal={arXiv preprint arXiv:2411.11581},
  year={2024}
}

@article{piao2025agentsociety,
  title={Agentsociety: Large-scale simulation of llm-driven generative agents advances understanding of human behaviors and society},
  author={Piao, Jinghua and Yan, Yuwei and Zhang, Jun and Li, Nian and Yan, Junbo and Lan, Xiaochong and Lu, Zhihong and Zheng, Zhiheng and Wang, Jing Yi and Zhou, Di and others},
  year={2025}
}

@article{park2024generative1000,
  title={Generative agent simulations of 1,000 people},
  author={Park, Joon Sung and Zou, Carolyn Q and Shaw, Aaron and Hill, Benjamin Mako and Cai, Carrie and Morris, Meredith Ringel and Willer, Robb and Liang, Percy and Bernstein, Michael S},
  journal={arXiv preprint arXiv:2411.10109},
  volume={52},
  year={2024}
}

@article{wang2026agentopia,
  title={Agentopia: Long-Term Life Simulation and Learning in Agent Societies},
  author={Wang, Xintao and Zheng, Sirui and Wu, Hongqiu and Li, Weiyuan and Huang, Jen-tse and Zhu, Minghao and Zu, Can and Deng, Qi and Wang, Jiawei and He, Qianyu and others},
  journal={arXiv preprint arXiv:2606.07513},
  year={2026}
}

@inproceedings{ji2026graphia,
  title={GRAPHIA: Harnessing Social Graph Data to Enhance LLM-Based Social Simulation},
  author={Ji, Jiarui and Zhang, Zehua and Wei, Zhewei and Tong, Bin and Wang, Guan and Zheng, Bo},
  booktitle={Proceedings of the 64th Annual Meeting of the Association for Computational Linguistics (Volume 1: Long Papers)},
  pages={7103--7128},
  year={2026}
}

@article{zhou2026odyssim,
  title={OdysSim: Building Foundation Models for Human Behavior Simulation},
  author={Zhou, Xuhui and Sun, Weiwei and Du, Weihua and Liu, Jiarui and Sun, Haojia and Ma, Qianou and Wu, Tongshuang and Yang, Yiming and Sap, Maarten},
  journal={arXiv preprint arXiv:2606.14199},
  year={2026}
}

@article{zhang2025socioverse,
  title={Socioverse: A world model for social simulation powered by llm agents and a pool of 10 million real-world users},
  author={Zhang, Xinnong and Lin, Jiayu and Mou, Xinyi and Yang, Shiyue and Liu, Xiawei and Sun, Libo and Lyu, Hanjia and Yang, Yihang and Qi, Weihong and Chen, Yue and others},
  journal={arXiv preprint arXiv:2504.10157},
  year={2025}
}

@article{wang2026ai,
  title={AI-generated Images Challenge Visual Trust in High-risk Scenarios},
  author={Wang, Yi-Zhi and Xiao, Yichen and Yue, Linan and Gao, Weibo and Du, Yichao and Fang, Pengfei and Di, Shimin and Zhang, Min-Ling},
  journal={arXiv preprint arXiv:2607.22745},
  year={2026}
}

@article{gong2026guided,
  title={Guided by Trajectories: Repairing and Rewarding Tool-Use Trajectories for Tool-Integrated Reasoning},
  author={Gong, Siyu and Yue, Linan and Gao, Weibo and Yao, Fangzhou and Di, Shimin and Feng, Lei and Zhang, Min-Ling},
  journal={arXiv preprint arXiv:2601.23032},
  year={2026}
}

@article{202609.0665,
	doi = {10.20944/preprints202609.0665.v2},
	url = {https://doi.org/10.20944/preprints202609.0665.v2},
	year = 2026,
	month = {September},
	publisher = {Preprints},
	author = {Zheng Zhang and Zhixiang Guo and Yuan Si and Siyuan Liang and Shunyu Liu and Weibo Gao and Song Wang and Leszek Rutkowski and Giuseppe Valenzise and Ming-Hsuan Yang and Lin William Cong and M. Jamal Deen and Sally Cripps and Dacheng Tao},
	title = {A Survey of World Model Benchmarks},
	journal = {Preprints}
}
\newpage
\end{document}